\documentclass[conf]{new-aiaa}
\usepackage[utf8]{inputenc}

\usepackage[version=4]{mhchem}
\usepackage{siunitx}
\usepackage{longtable,tabularx}
\usepackage{amsmath,amsfonts}

\usepackage[ruled,vlined,linesnumbered]{algorithm2e}
\usepackage{algpseudocode}
\usepackage{graphicx}
\usepackage[font={footnotesize}]{caption}
\usepackage{subcaption}
\usepackage{textcomp}
\usepackage[dvipsnames]{xcolor}
\usepackage{setspace}
\usepackage{enumerate}
\usepackage{comment}

\newcommand{\R}{{\mathbb{R}}}
\def\BibTeX{{\rm B\kern-.05em{\sc i\kern-.025em b}\kern-.08em
    T\kern-.1667em\lower.7ex\hbox{E}\kern-.125emX}}

\newcommand{\tf}{{t_{\rm f}}}

\usepackage{titlesec}   

\titlespacing*{\section}{0pt}{3pt}{10pt}
\titlespacing*{\subsection}{0pt}{3pt}{5pt}

\title{Kinodynamic FMT* with Dimensionality Reduction Heuristics and Neural Network Controllers}

\author{Dongliang Zheng\footnote{Graduate Student, School of Aerospace Engineering.} and 
Panagiotis Tsiotras\footnote{David and Andrew Lewis Chair Professor, School of Aerospace Engineering and Institute for Robotics and Intelligent Machines; 
AIAA Fellow.}}
\affil{Georgia Institute of Technology, Atlanta, GA, 30332-0150}

\begin{document}

\maketitle

\begin{abstract}
This paper proposes a new sampling-based kinodynamic motion planning algorithm, called FMT*PFF, for nonlinear systems. 
It exploits the novel idea of dimensionality reduction using partial-final-state-free (PFF) optimal controllers.
With the proposed dimensionality reduction heuristic, the search space is restricted within a subspace, thus faster convergence is achieved compared to a regular kinodynamic FMT*.
The dimensionality reduction heuristic can be viewed as a sampling strategy and asymptotic optimality is preserved when combined with uniform full-state sampling.
Another feature of FMT*PFF is the ability to deal with a steering function with inexact steering, which is vital when using learning-based steering functions.
Learning-based methods allow us to solve the steering problem for nonlinear systems efficiently. However, learning-based methods often fail to reach the exact goal state. 
For nonlinear systems, we train a neural network controller using supervised learning to generate the steering commands.
We show that FMT*PFF with a learning-based steering function is efficient and generates dynamically feasible motion plans.
We compare our algorithm with previous algorithms and show superior performance in various simulations.
\end{abstract}

\section{Introduction}\label{sec:intro}

Motion planning, as a fundamental component of robot autonomy, has been studied extensively in the last three decades to increase its efficiency and capability.
Efficiency means faster convergence to better solutions.
Efficient planning algorithms are crucial for robots with limited computation power and for replanning in changing environments.
Capability means dealing with more complicated planning problems. 
High-dimensional state space, nonlinear system dynamics, and cluttered environments with nonconvex obstacles still pose great challenges for efficient kinodynamic planning despite recent advances.

Sampling-based motion planning algorithms, such as PRM \cite{Kavraki1996Prob} and RRT \cite{lavalle2001randomized}, have been developed to solve planning problems in high-dimensional continuous state spaces by incrementally building a graph/tree through the search space.
The optimal sampling-based planning algorithm RRT*~\cite{Karaman2011Sampling} almost surely converges asymptotically to the optimal solution.
RRT* is well-suited for planning in high-dimensional spaces and obstacle-rich environments.
Many applications of RRT* have been studied in recent years~\cite{gammell2021asymptotically}.

For motion planning for dynamical systems, sampling-based optimal kinodynamic planning algorithms (SBKMP) such as Kinodynamic RRT* \cite{Karaman2010Optimal} and Kinodynamic FMT* \cite{Schmerling2015Optimal} have been developed to consider differential constraints.
SBKMP requires any two points sampled in the planning space to be connected with an optimal trajectory.
For robots with differential constraints, the optimal trajectory between two states is obtained by solving a two-point boundary value problem (TPBVP), which is a non-trivial undertaking for complex nonlinear systems. 
The solution to this local TPBVP is also referred to as the steering function. Simulation-based methods such as SST \cite{Li2016Asymptotically} avoid solving the TPBVP by using random control sampling and simulation. However, without the local optimal edges (trajectories) provided by the steering function, the convergence of SST to a good solution is slow.

Solving TPBVPs efficiently for nonlinear systems is one of the bottlenecks of kinodynamic RRT*. 
Thus, researchers have looked into more efficient ways to solve these TPBVPs.
A steering function based on LQR is used in~\cite{Perez2012LQR}.
A fixed-final-state free-final-time controller for linear systems that optimally connects any pair of states is introduced in~\cite{Dustin2013Kinodynamic}. 
While an analytical solution of the TBPVP for linear systems is available, considering general nonlinear system dynamics is difficult. 
Learning-based methods have the potential for solving TPBVP efficiently.
To deal with nonlinear dynamics, steering functions based on supervised learning and reinforcement learning are developed in \cite{Wolfslag2018RRT-CoLearn} and \cite{Chiang2019RLRRT} respectively, and integrated with a kinodynamic RRT* algorithm. In \cite{zheng2021sampling, zheng2021Near}, a goal-conditioned state-feedback neural network controller for nonlinear systems is trained and used for solving the TPBVPs in RRT*.

Another limitation of RRT* is the slow convergence rate of the solution to the optimal one, which is especially evident for the kinodynamic planning case where the sampling space is not just the configuration space but the full state space.
Heuristic and informed sampling methods have been developed to improve the convergence rate.
Informed RRT* \cite{Gammell2014Informed} focuses sampling to an informed subset that could potentially provide a better solution. 
Exploiting the benefit of ordered search, FMT* \cite{janson2015fast} and BIT* \cite{gammell2020batch} are shown to find better solutions faster than RRT*. 
However, most of these methods only consider the geometric planning problem. 
Existing work on heuristics for improving the convergence of kinodynamic motion planning is rather limited~\cite{paden2017verification,yi2018generalizing}.
In our previous work \cite{Zheng2021Accelerating}, the Kino-RRT* with a dimensionality reduction heuristic is developed, in which the heuristic is obtained by solving a partial final-state free (PFF) optimal control problem. 
Instead of sampling the full state space, Kino-RRT* only samples part of the state space while the rest of the states are selected by the PFF optimal controller. By sampling in the reduced state space and utilizing the PFF optimal controller, Kino-RRT* shows faster convergence. An analytical solution for the PFF optimal control problem for linear systems is also derived in \cite{Zheng2021Accelerating}.

In this paper, we propose the FMT*PFF, which is built on our previous works \cite{zheng2021sampling} and \cite{Zheng2021Accelerating}. We extend the dimensionality reduction heuristic to learning-based planners and train neural network controllers to solve the PFF optimal control problem.
Compared to \cite{zheng2021sampling}, the dimensionality reduction heuristic is used in FMT*PFF to improve the convergence rate of the algorithm. Also, training a PFF neural network controller is simpler than the set-to-set controller in \cite{zheng2021sampling}.
Compared to \cite{Zheng2021Accelerating}, solving PFF using supervised learning allows us to deal with nonlinear system dynamics efficiently.
Furthermore, while \cite{zheng2021sampling} and \cite{Zheng2021Accelerating} are based on the RRT* algorithm, FMT*PFF is based on the FMT* algorithm to benefit from ordered search.

The contributions of the paper are:
\begin{itemize}
  \item A dimensionality reduction heuristic for accelerating sampling-based kinodynamic planning for nonlinear systems. 
  \item A neural network controller for solving the partial final-state free (PFF) optimal control problem. The proposed PFF neural network controller is used as the steering function in sampling-based kinodynamic motion planning algorithms.
  \item The FMT*PFF algorithm is developed for planning with learning-based steering functions that cannot achieve exact steering. 
  \item Extensive simulations and comparison with previous methods show better performance with our method.  
\end{itemize}

\section{Preliminaries}\label{sec:Preliminaries}

\subsection{Problem Statement}

The optimal kinodynamic motion planning problem is given by the following optimal control problem (OCP), 
\begin{subequations}
\begin{align}
\min_{u, t_{\mathrm{f}}} \ \ J &= \int_{0}^{t_{\mathrm{f}}} c(x,u) \, \mathrm{d}\tau, \label{eq:OCP1}\\
\mathrm{s.t.} \ \ \ \dot{x} &= f(x,u), \\
x(0) &=x_s, \ x(t_{\mathrm{f}})=x_g, \\
u & \in \mathcal{U}, \ x \in \mathcal{X}_{\mathrm{free}}, \ \forall t\in [0, t_{\mathrm{f}}].
\label{eq:OCP}
\end{align}
\end{subequations}
where $x \in \mathcal X \subset \R^{n_x}$ is the state, $u \in \mathcal U \subset \R^{n_u}$ is the control input, $x_s$ and $x_g$ are the initial state and goal state, respectively. The free space is denoted by $\mathcal X_{\mathrm{free}} \subset \mathcal{X}$, where at each $x \in \mathcal X_{\mathrm{free}}$, the system does not collide with any obstacles in the environment. Finally, $c(x, u)$ is a cost function that we aim to minimize.

The goal of the optimal kinodynamic motion planning problem is to find a control trajectory $u(t)$, $t \in [0,t_{\mathrm{f}}]$, such that the solution state trajectory $x(t)$ is obstacle-free, reaches the goal state, and minimizes a cost.

\subsection{Partial-Final-State-Free Optimal Controller}

Sampling-based motion planning algorithms such as RRT* and FMT* solve the problem (\ref{eq:OCP1})-(\ref{eq:OCP}) by growing a tree.
The state space is approximated by random samples. 
The transition between samples is achieved using optimal steering functions. 
The sampled nodes and the connections between nodes define a graph.
A trajectory tree is obtained by searching over this graph.

In kinodynamic RRT* and FMT*, the edge between two sampled states, $x_a$ and $x_b$, is constructed using a steering function which is the solution of a TPBVP given by

\begin{subequations}
\begin{align}
\min_{u, t_{\mathrm{f}}} \ \ J &= \int_{0}^{t_{\mathrm{f}}} c(x,u) \, \mathrm{d}\tau, \label{eq:TPBVP1} \\
\mathrm{s.t.} \ \ \ \dot{x} &= f(x,u), \label{eq:TPBVP2}\\
 x(0) &=x_a, \ x(t_{\mathrm{f}})=x_b, \label{eq:TPBVP3}\\
 u & \in \mathcal{U}, \ \forall t\in [0, t_{\mathrm{f}}].
\label{eq:TPBVP}
\end{align}
\end{subequations}
Note that the obstacle constraint in (\ref{eq:OCP}) is removed in TPBVP.

In our proposed FMT*PFF, instead of sampling the full state $x$ from the full state space $\mathcal{X}$, we sample a partial state $\bar{x}$ from a state space of reduced dimensionality $\bar{\mathcal{X}}$. 
Let $x = [x_1^\top \ x_2^\top]^\top$, where $x_1 \in \mathbb{R}^{n_1}$, $x_2 \in \mathbb{R}^{n_2}$, and $n_1 + n_2 = n_x$. 
We introduce the partial-final-state-free (PFF) optimal control problem as follows
\begin{subequations}
\begin{align}
\min_{u, t_{\mathrm{f}}} \ \  J &= \int_{0}^{t_{\mathrm{f}}} c(x,u) \, \mathrm{d}\tau, \label{eq:PFFOCP1}\\
\mathrm{s.t.} \ \ \  \dot{x} &= f(x,u), \label{eq:PFFOCP2}\\
 x(0) &=x_a, \ x_1(t_{\mathrm{f}})=\bar{x}_c, \label{eq:PFFOCP3}\\
u & \in \mathcal{U}, \ \forall t\in [0, t_{\mathrm{f}}].
\label{eq:PFFOCP}
\end{align}
\end{subequations}
Compared to (\ref{eq:TPBVP3}), instead of fixing the state $x(\tf)$, only $x_1(\tf)$ is fixed and $x_2(\tf)$ is free in (\ref{eq:PFFOCP3}).

Note that, after solving the problem (\ref{eq:PFFOCP1})-(\ref{eq:PFFOCP}), we obtain the full state trajectory which includes the full final state $x(t_{\mathrm{f}})$. 
Thus, the PFF optimal controller chooses the remaining free final state to minimize the cost. 
If we set this full final state $x(t_{\mathrm{f}})$ as the terminal state in the terminal constraint in problem (\ref{eq:TPBVP1})-(\ref{eq:TPBVP}) and solve the problem (\ref{eq:TPBVP1})-(\ref{eq:TPBVP}), we will get the same trajectory as in problem (\ref{eq:PFFOCP1})-(\ref{eq:PFFOCP}).
Therefore, the partial state sampling and the PFF optimal controller work as an intelligent heuristic for state-space sampling.

For linear systems and quadratic cost functions, the analytical solution of the problem (\ref{eq:PFFOCP1})-(\ref{eq:PFFOCP}) is derived in \cite{Zheng2021Accelerating}.
For general nonlinear systems and cost functions, solving (\ref{eq:PFFOCP1})-(\ref{eq:PFFOCP}) efficiently is the bottleneck for sampling-based kinodynamic planning algorithms.
Methods that integrate a numerical solver within a sampling-based planner to solve (\ref{eq:TPBVP1})-(\ref{eq:TPBVP}) have been studied \cite{Xie2015Towards}. However, the long computation time makes them unacceptable for practical use. 
In this paper, we use the PFF optimal controller as the steering function.
We train a neural network controller to solve the PFF optimal control problem.
We show that the proposed FMT*PFF algorithm with a neural network controller is efficient and generates dynamically feasible motion plans.

\section{The FMT*PFF Algorithm} \label{sec:FMT*}

The main differences between FMT*PFF and the original kinodynamic FMT* are: 1) state sampling in a reduced state space; 2) use of the PFF optimal controller as the steering function. In our case, the PFF optimal controller is approximated using a neural network.

\begin{figure}
    \centering
    \begin{subfigure}[b]{0.24\columnwidth}
         \centering
         \includegraphics[width=0.9\columnwidth]{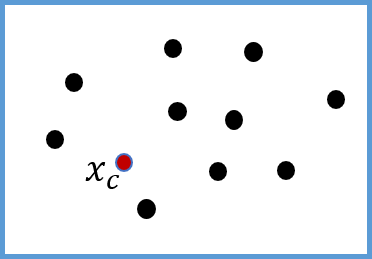}   
         \caption{}
     \end{subfigure}
     \begin{subfigure}[b]{0.24\columnwidth}
         \centering
         \includegraphics[width=0.9\columnwidth]{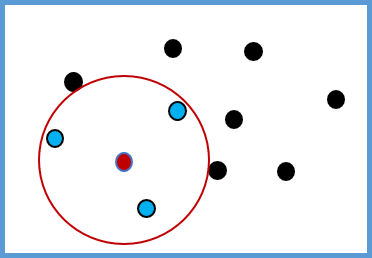}   
         \caption{}
     \end{subfigure}
     \begin{subfigure}[b]{0.24\columnwidth}
         \centering
         \includegraphics[width=0.9\columnwidth]{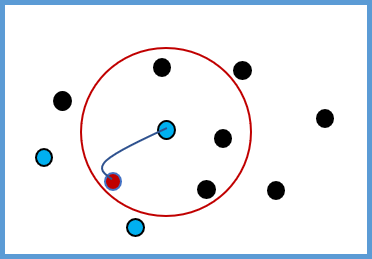}   
         \caption{}
     \end{subfigure}
     \begin{subfigure}[b]{0.24\columnwidth}
         \centering
         \includegraphics[width=0.9\columnwidth]{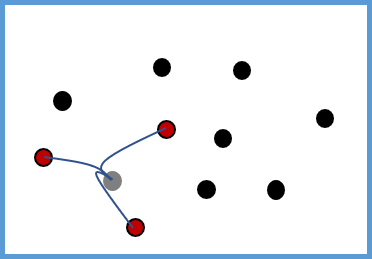}   
         \caption{}
     \end{subfigure}
     \begin{subfigure}[b]{0.24\columnwidth}
         \centering
         \includegraphics[width=0.9\columnwidth]{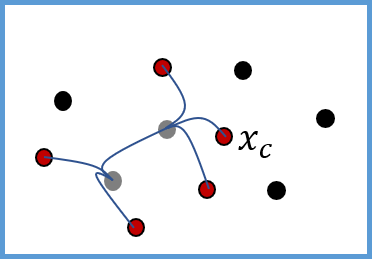}   
         \caption{}
     \end{subfigure}
     \begin{subfigure}[b]{0.24\columnwidth}
         \centering
         \includegraphics[width=0.9\columnwidth]{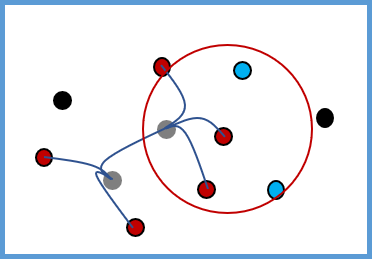}   
         \caption{}
     \end{subfigure}
     \begin{subfigure}[b]{0.24\columnwidth}
         \centering
         \includegraphics[width=0.9\columnwidth]{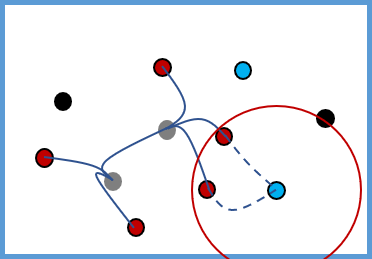}   
         \caption{}
     \end{subfigure}
     \begin{subfigure}[b]{0.24\columnwidth}
         \centering
         \includegraphics[width=0.9\columnwidth]{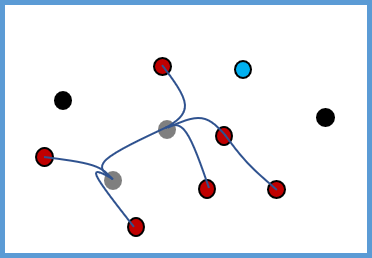}   
         \caption{}
     \end{subfigure}
        \caption{An illustration of the FMT*PFF algorithm. (a) $m$ new samples, shown as black dots, are sampled in the reduced state space using $\mathsf{SamplePFF}$. The starting state $x_s$ ( red dot) is added to the tree. Since it is the only state in the tree (and the set $X_{\mathrm{open}}$), $x_c$ is set to be $x_s$. (b) The neighboring states, shown in blue, are found. (c) Find the best parent in the tree for all neighboring states. The partial final-state free (PFF) controller is used for steering. (d) Completed One iteration. (e) At some iteration of the algorithm, the state $x_c$ with the minimum cost-to-come in $X_{\mathrm{open}}$ is selected. (f) The neighboring states, shown in blue, are found. (g) Find the best parent in the tree for a neighboring state using $\mathsf{SteerPFF}$. (h) The best parent is found. The new state $x_{\mathrm{new}}$ is added to the tree.}
        \label{FMT*PFFillstration}
\end{figure}

\IncMargin{.5em}
\begin{algorithm}
\caption{FMT*PFF}
\label{alg:FMT*PFF}
$v_s.x \leftarrow x_s$; $v_s.\bar{x} \leftarrow \bar{x}_{s}$\;
$V \leftarrow \{v_s\}$; $E \leftarrow \emptyset$; $G \leftarrow (V,E)$\;
$\bar{X}_\mathrm{unvisited} \leftarrow \{\bar{x}_g\} \cup \mathsf{SamplePFF}(m)$\;
$X_\mathrm{open} \leftarrow \{x_s\}$\;
$x_c \leftarrow x_s$; $\bar{x}_c \leftarrow \bar{x}_s$\;
\While{$\bar{x}_c \neq \bar{x}_g$}
{   
    $X_\mathrm{open,new} \leftarrow \emptyset$\;
    $\bar{X}_\mathrm{near} \leftarrow \mathsf{Near}(x_c, \bar{X}_\mathrm{unvisited})$\;
    \ForEach{$\bar{x}_\mathrm{near} \in \bar{X}_\mathrm{near}$}
    {
        $X_\mathrm{near} \leftarrow \mathsf{Near}(X_\mathrm{open}, \bar{x}_\mathrm{near})$\;
        $x_\mathrm{parent} = \underset{x \in X_\mathrm{near}}{\mathrm{argmin}}(\mathsf{cost}(x) + \mathsf{SegCost}(x, \bar{x}_\mathrm{near}))$\;
        $(\tau, x_\mathrm{new}) \leftarrow \mathsf{SteerPFF}(x_\mathrm{parent}, \bar{x}_\mathrm{near})$\;
        \If{$\mathsf{CollisionFree}(\tau)$}
        {
            $v.x \leftarrow x_\mathrm{new}$; $v.\bar{x} \leftarrow \bar{x}_\mathrm{near}$\;
            $V \leftarrow  V \cup \{v\}$\;
            $E \leftarrow E \cup \{ (v(x_\mathrm{parent}), v)\}$\;
            $\bar{X}_\mathrm{unvisited} \leftarrow \bar{X}_\mathrm{unvisited} \setminus \bar{x}_\mathrm{near}$\;
            $X_\mathrm{open,new} \leftarrow X_\mathrm{open,new} \cup \{x_\mathrm{new}\}$\;
        }
    }
    $X_\mathrm{open} \leftarrow X_\mathrm{open} \cup X_\mathrm{open,new} \setminus \{x_c\}$\;
    \If{$X_\mathrm{open} = \emptyset$}
    {
        \KwRet $\mathsf{Failure}$;
    }
    $x_c = \underset{x \in X_\mathrm{open}}{\mathrm{argmin}}(\mathsf{cost}(x))$\;
    $\bar{x}_c \leftarrow v(x_c).\bar{x}$
}
\KwRet $G$;
\end{algorithm}
\DecMargin{.5em}

The FMT*PFF algorithm is given by Algorithm \ref{alg:FMT*PFF} and a graphical illustration is given in Figure~\ref{FMT*PFFillstration}.
We use $\bar{x}$ to represent a partial state in the reduced state space and $\bar{X}$ to represent the partial state set. 
We use $x$ to represent a (full)) state in the (full) state space, and $X$ to represent the set of (full) states.
Some primitive procedures are given as follows.\\
\textbf{Sampling:} The sampling procedure $\mathsf{SamplePFF}(m)$ randomly samples $m$ partial states in the reduced state space. The sampled partial states are collision-free in the corresponding reduced state space.
For example, for a robot whose state space includes the position space and the velocity space, $\mathsf{SamplePFF}$ samples positions of the robot that are collision-free.\\ 
\textbf{Near Nodes:} The function $\mathsf{Near}(x,\bar{X})$ returns all the partial states in $\bar{X}$ that are contained in a ball of radius $r$ centered at $x$. The function $\mathsf{Near}(X,\bar{x})$ returns all the states in $X$ that are contained in a ball of radius $r$ centered at $\bar{x}$.
One simple implementation of the distance function is the Euclidean distance in the partial state space.\\ 
\textbf{Collision Checking:} The function $\mathsf{CollisionFree}(\tau)$ takes a trajectory $\tau$ (an edge segment) as an input and returns true if and only if $\tau$ lies entirely in the collision-free space. \\
\textbf{Cost:} The procedure $\mathsf{Cost}(x)$ returns the cost-to-come from the root node to $x$. \\
\textbf{Segment Cost:} The procedure $\mathsf{SegCost}(x_i,\bar{x}_j)$ returns the cost to go from $x_i$ to $\bar{x}_j$. 
This cost is obtained by solving the PFF optimal control problem with boundary conditions $x_i$ and $\bar{x}_j$. We can train a neural network to predict this edge cost.\\
\textbf{Steering:} The procedure $\mathsf{SteerPFF}(x_i,\bar{x}_j)$ solves the TPBVP using the PFF optimal controller, and it returns a trajectory $\tau$ that starts from $x_i$ and ends at $\bar{x}_j$.\\

In Algorithm \ref{alg:FMT*PFF}, every vertex $v$ is associated with a state $v.x$ and the corresponding partial state $v.\bar{x}$. 
We initialize the tree in Line 1-2.
In Line 3, the $m$ partial states and the goal are added to the unvisited set $\bar{X}_\mathrm{unvisited}$, which is a set of partial states that have not been added to the tree.
$X_\mathrm{open}$ is the set of states that have already been added to the tree. They are the frontier nodes of the tree that will be extended next. 
$x_c$ is the state in $X_\mathrm{open}$ that has the minimum cost-to-come (Line 22) and $\bar{x}_c$ is the partial state associated with $x_c$. These are initialized in Line 5.
If $\bar{x}_c = \bar{x}_g$, the solution has been found. Otherwise, we try to extend the tree.

In Line 8, the neighboring partial states of $x_c$ in $\bar{X}_\mathrm{unvisited}$, $\bar{X}_\mathrm{near}$, are found. 
For every $\bar{x}_\mathrm{near}$ in $\bar{X}_\mathrm{near}$, we try to find its best parent in $X_\mathrm{open}$ (Line 9-18). 
In Line 10, the neighboring state of $\bar{x}_\mathrm{near}$ in $X_\mathrm{open}$, $X_\mathrm{near}$, are found. 
The best parent for $\bar{x}_\mathrm{near}$ that results in the minimum cost-to-come for $\bar{x}_\mathrm{near}$ is obtained in Line 11. The trajectory $\tau$ and the full state $x_{\mathrm{new}}$ are obtained in Line 12 using the PFF steering function.
If $\tau$ is collision-free, the new vertex $v$ and new edge are added to the tree, and the sets $\bar{X}_\mathrm{unvisited}$ and $X_\mathrm{open,new}$ are updated (Line 13-18).
$v(x_\mathrm{parent})$ denotes the vertex associated with $x_\mathrm{parent}$ and $v(x_c).\bar{x}$ denotes the partial state associated with $x_c$.

\subsection{A Neural Network Controller for PFF Optimal Control}

The proposed FMT*PFF can directly work with the analytical solution of the PFF optimal control problem, a numerical solver for the PFF optimal control problem, and a learning-based steering function for the PFF optimal control problem.
In this section, we describe training the neural network controller for PFF optimal control.

We first generate the training data. 
For a dynamical system, we use numerical optimization solvers~\cite{Patterson2014GPOPS-II} to solve the PFF optimal control problem given by (\ref{eq:PFFOCP1})-(\ref{eq:PFFOCP}) offline.
Since we are interested in steering the system from different initial states to the partial final state, we generate optimal trajectories with different initial states. 
The initial states are uniformly sampled from an initial state set. 
We choose the partial final state to be the position, and let the remaining state (heading angle, velocity, etc.) be free. 

We utilize the translational invariant of the trajectories.
If the goal position is not the origin, we can translate the goal position to the origin and translate the position of the starting state accordingly. After solving the optimal control problem with the translated boundary condition, we can get the trajectory of the original problem by translating the solution trajectory back. 
Therefore, we choose the goal position to be at the origin (zero). 
Translational invariance regarding the position is common for systems such as double integrator, car, and UAVs.

Note that we only need to steer the system to nearby states given by the $\mathsf{Near}$ function in FMT*PFF. 
Thus, a 'local' PFF optimal controller is sufficient. 
This makes the learning-based controller using a neural network well-suited for this task since we can only generate finite training data. 
When using the neural network for prediction, it is important to make sure we are using it for interpolation instead of extrapolation.
Extrapolation will diminish the prediction accuracy.
For this purpose, the offline training data should cover the neighborhood used in the $\mathsf{Near}$ function.

Each trajectory returned by the numerical solver contains a sequence of control inputs and a sequence of states indexed by time.
We combine all the data points from all offline trajectories to form the final training dataset.
Each data point is a tuple $(x_i,u_i)$. 
The goal of the neural network is to mimic the structure of the optimal controller. 
The neural network controller, $\mathsf{NNController}: x_i \rightarrow u_i$, is a state feedback controller mapping from the current state $x_i$ to the current control $u_i$ to be applied.

After training the neural network controller, it is used for online control and online trajectory generation. The $\mathsf{SteerPFF}$ in Algorithm~\ref{alg:FMT*PFF} is obtained by applying the neural network controller to get a trajectory. 
Given a novel initial state, which should be inside the initial state set used in training data generation, we repetitively apply the neural network control commands and simulate the system dynamics to obtain the trajectory that steers the system to the goal state, where the goal position is the origin. 
We also train a cost-to-go neural network that predicts the segment cost, $\mathsf{SegCost}: x_1 \rightarrow J$, where $x_1$ is the initial state, $J$ is the cost of the trajectory from $x_1$ to the goal returned by the numerical solver. 

\begin{figure}
    \centering
    \begin{subfigure}[b]{1\columnwidth}
         \centering
         \includegraphics[width=0.49\columnwidth]{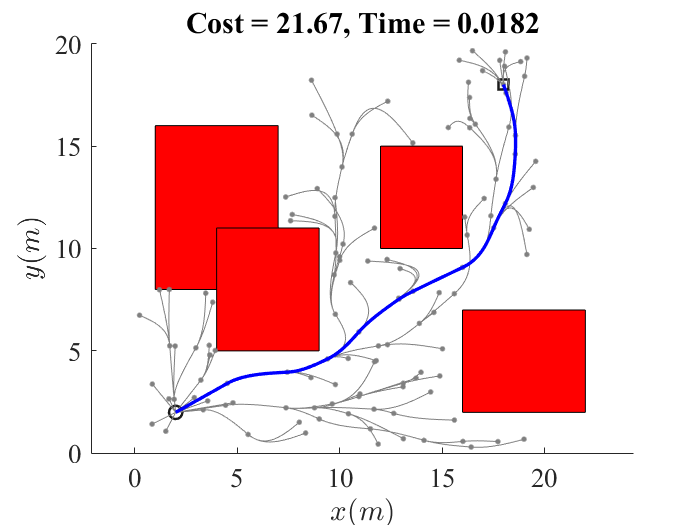}
         \includegraphics[width=0.49\columnwidth]{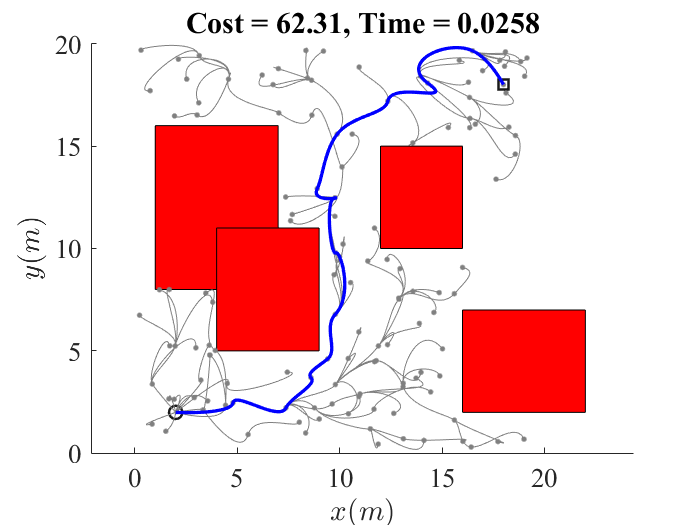}         
     \end{subfigure}
     \begin{subfigure}[b]{1\columnwidth}
         \centering
         \includegraphics[width=0.49\columnwidth]{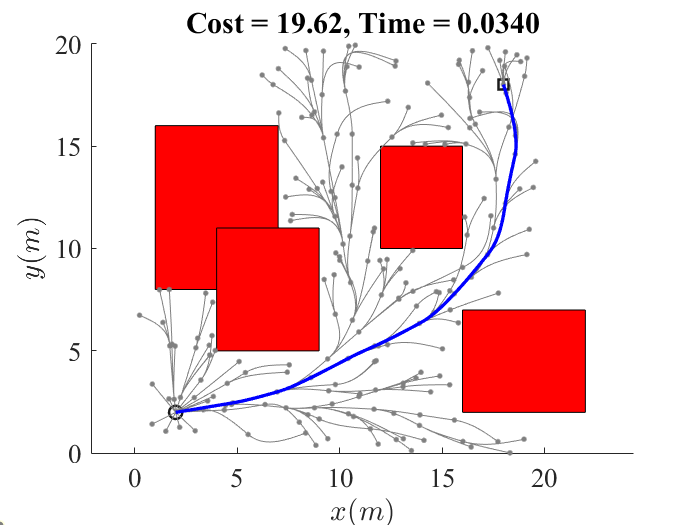}
         \includegraphics[width=0.49\columnwidth]{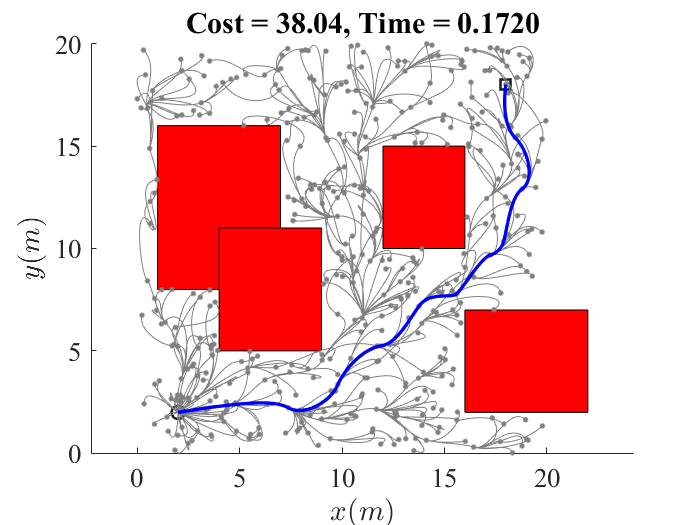}         
     \end{subfigure}
     \begin{subfigure}[b]{1\columnwidth}
         \centering
         \includegraphics[width=0.49\columnwidth]{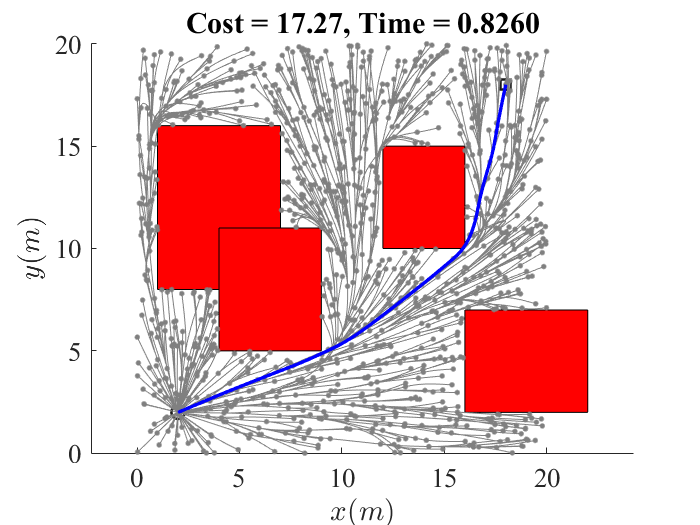}
         \includegraphics[width=0.49\columnwidth]{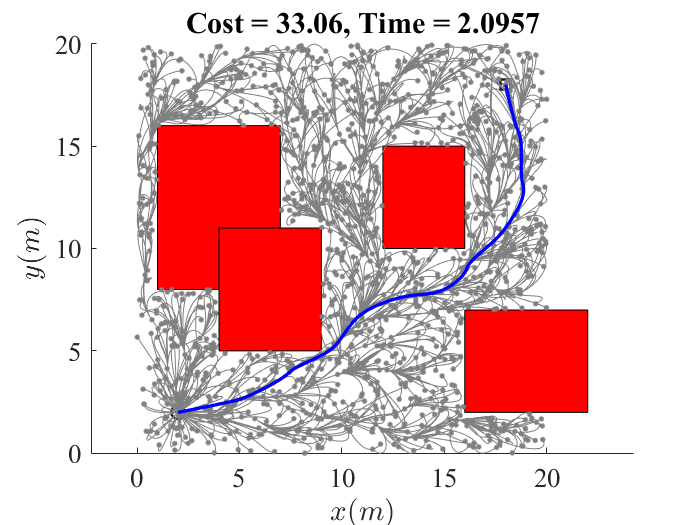}         
     \end{subfigure}
        \caption{FMT*PFF planning results. In each figure, the blue line is the solution from the current tree. The figures in the first column show the planning results of our proposed FMT*PFF algorithm with an increasing number of samples. The figures in the second column are the planning results of the original kinodynamic FMT*.}
        \label{TreeDIcomp}
\end{figure}

\section{Empirical Evaluation} \label{sec:sim}

\subsection{2D Double Integrator} \label{SubSec2DDI}

We first compare FMT*PFF with the kinodynamic FMT* \cite{Schmerling2015Optimal} and the Kino-RRT* \cite{Zheng2021Accelerating}.
For this purpose, we will consider linear systems and use the analytical solution for (\ref{eq:TPBVP1})-(\ref{eq:TPBVP}) and (\ref{eq:PFFOCP1})-(\ref{eq:PFFOCP}). 
The difference between FMT*PFF and the kinodynamic FMT* is that kinodynamic FMT* samples the full state-space and solves (\ref{eq:TPBVP1})-(\ref{eq:TPBVP}) for steering functions while FMF*PFF samples in the reduced state-space and solves (\ref{eq:PFFOCP1})-(\ref{eq:PFFOCP}) for steering functions. 
For comparison, we set the tuning parameters of the algorithms, such as the neighborhood radius $r$, to be the same. 
Both Kino-RRT* and FMT*PFF use the PFF optimal control, but they are based on RRT* and FMT*, respectively.

The state of the 2D double integrator is given by $x = [p^\top \ v^\top]^\top$, where $p = [x_1 \ x_2]^\top$ is the position and $v = [x_3 \ x_4]^\top$ is the velocity.
The control input is acceleration.
The system dynamics is given by $\dot{x} = A x + B u$, where 
\begin{equation*}
    A = \begin{bmatrix} 0 & I_2 \\ 0 & 0 \end{bmatrix}, \quad
    B = \begin{bmatrix} 0 \\ I_2 \end{bmatrix}.
\end{equation*}
The cost function $c(x,u) = 1 + u^\top R u$, where $R = I_2$.

The position is uniformly sampled within the boundary of the environment.
The free final state of the PFF controller is the velocity.
Thus, FMT*PFF only samples the position space.
For the kinodynamic FMT* algorithm, the velocity is uniformly sampled in $v \in [-2, 2]^2 \ \mathrm{m/s^2}$.  
Note that a larger interval for the velocity essentially requires searching in a larger state space, which will result in slower convergence.
However, if the sampling velocity interval is too small, the search is confined to a small state space that may not contain the optimal solution.  

The planning results of the FMT*PFF algorithm and the Kinodynamic FMT* algorithm are given in Figure~\ref{TreeDIcomp}. By sampling in the reduced state space and using a PFF optimal controller, FMT*PFF reduced the dimensionality of the planning problem. Thus, FMT*PPF finds a better trajectory from the beginning and continues to find better solutions given the same amount of planning time.
For Kinodynamic FMT*, the probability of sampling good velocities to decrease the cost is low.

\begin{figure}
    \centering
    \includegraphics[width=0.5\columnwidth]{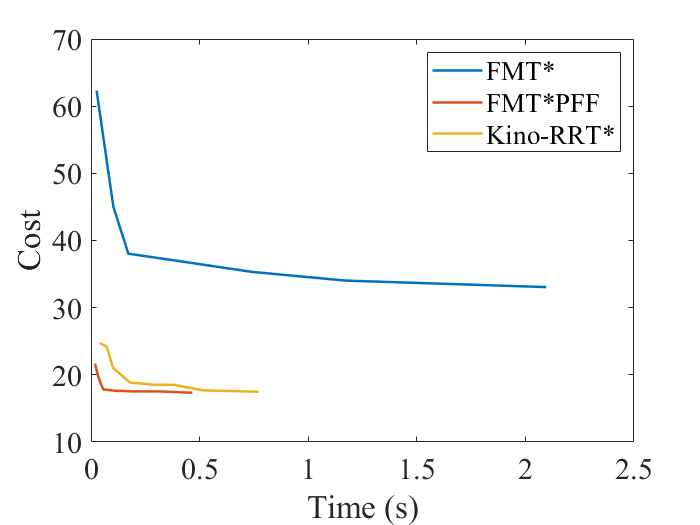}
        \caption{Comparison of FMT*, FMT*PFF,  and Kino-RRT* algorithm. Compared with FMT*, the dimensionality reduction heuristic in Kino-RRT* and FMT*PFF improve their performance significantly. FMT*PFF also performs better than Kino-RRT* due to its ordered search.}
        \label{CostDIcomp}
\end{figure}

The solution cost vs planning time comparison is given in Figure~\ref{CostDIcomp}.
We can see from Figure~\ref{CostDIcomp}, FMT*PFF also has better convergence performance compared to Kino-RRT*. 
This is because FMT*PFF uses ordered search while Kino-RRT* uses unordered random search.
One drawback of FMT*PFF is that it is not an anytime algorithm.
Still, the benefit of FMT*PFF becomes more clear when planning with learning-based steering functions that cannot achieve exact steering, since FMT*PFF does not use a rewiring procedure.

\begin{figure}
    \centering
    \begin{subfigure}[b]{1\columnwidth}
         \centering
         \includegraphics[width=0.5\columnwidth]{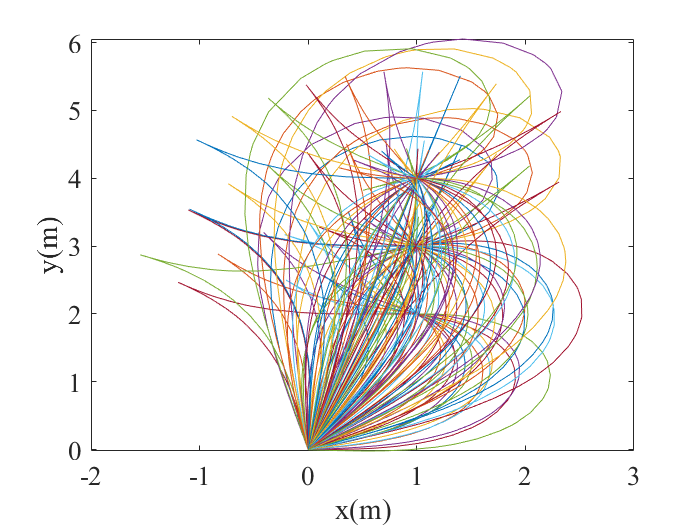}   
     \end{subfigure}
     \begin{subfigure}[b]{1\columnwidth}
         \centering
         \includegraphics[width=0.5\columnwidth]{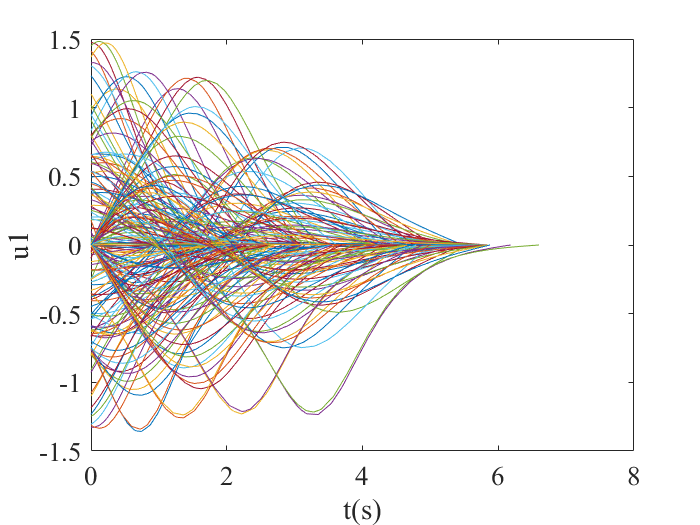}        
     \end{subfigure}
     \begin{subfigure}[b]{1\columnwidth}
         \centering
         \includegraphics[width=0.5\columnwidth]{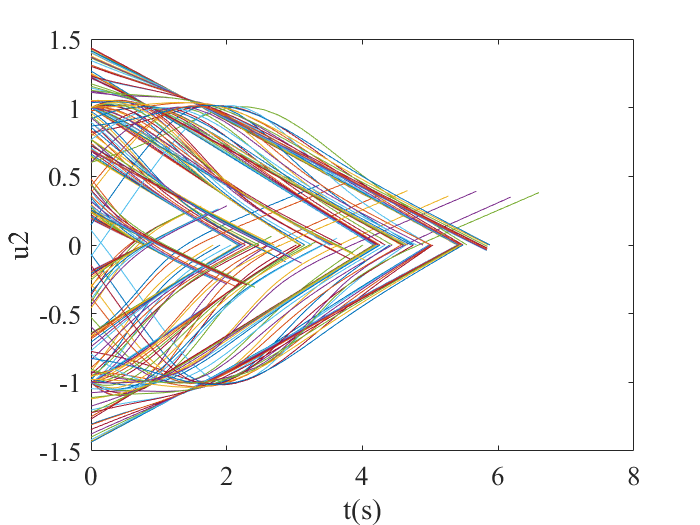}        
     \end{subfigure}
        \caption{Offline generated trajectories for neural network training. The first figure gives the trajectories of the position. The last two figures show the control trajectories.}
        \label{OfflineTraj}
\end{figure}

\subsection{A Simple Car Model} \label{SubSecCar}

The kinematic car model is given by
\begin{subequations}
\begin{align}
\dot{x} & = v\cos(\theta), \\
\dot{y} & = v\sin(\theta), \\
\dot{\theta} & = u_1, \\
\dot{v} & = u_2,
\end{align}
\end{subequations}
where $(x, y)$ is the position, $\theta$ is the heading angle, $v$ is the speed, $u_1$ and $u_2$ are the control inputs.

\begin{figure}
    \centering
    \includegraphics[width=0.5\columnwidth]{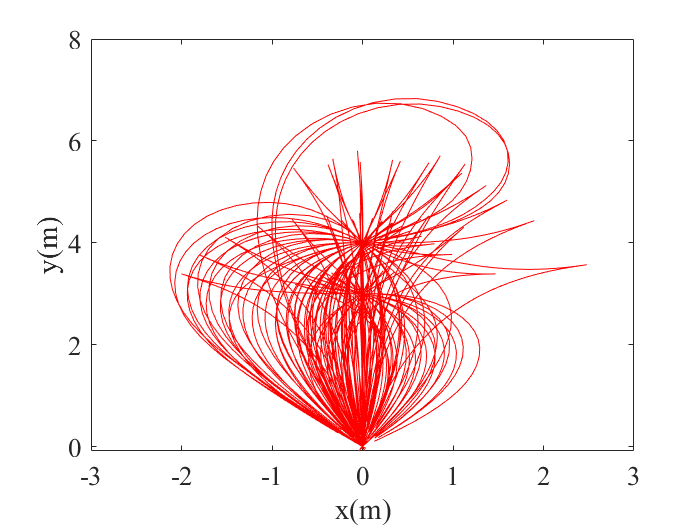}  
        \caption{Trajectories generated using the neural network controller. The initial states are sampled from the initial state set. The goal of the learning-based steering function is to steer the system to reach the goal position (the origin). By repeatedly applying the state-feedback neural network controller and forward simulating the system dynamics, the system reaches a neighborhood of the goal.}
        \label{TestTraj}
\end{figure}

We first generate offline training data by solving PFF optimal control problems using numerical optimization solvers.
The offline trajectory examples used for training are shown in Figure~\ref{OfflineTraj}. We sample initial states from an initial state set.
The sampling intervals of $x$, $y$, $\theta$, and $v$ are $x \in [-4, 4] \ \mathrm{m}$, $y \in [-4, 4] \ \mathrm{m}$, $\theta \in [-\pi, \pi] \ \mathrm{rad}$, and $v \in [-2, 2] \ \mathrm{m/s}$, respectively. 
The reduced state space is the position space and $\theta$ and $v$ are free states.
Thus, the goal state is $(x, y, \theta, v) = [0 \ 0 \ \mathrm{free} \ \mathrm{free}]$.
For this example, 10,400 trajectories are generated. 
State-action pairs from the trajectories were used for neural network training.

\begin{figure}[htb]
    \centering
    \begin{subfigure}{1\columnwidth}
         \centering
         \includegraphics[width=0.49\columnwidth]{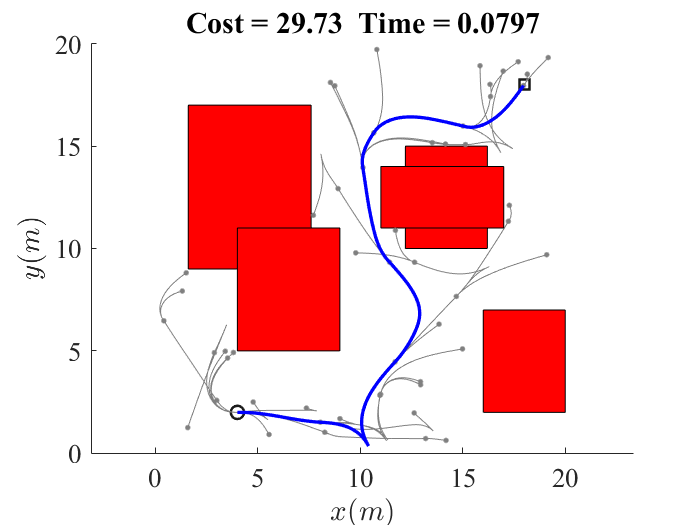}
         \includegraphics[width=0.49\columnwidth]{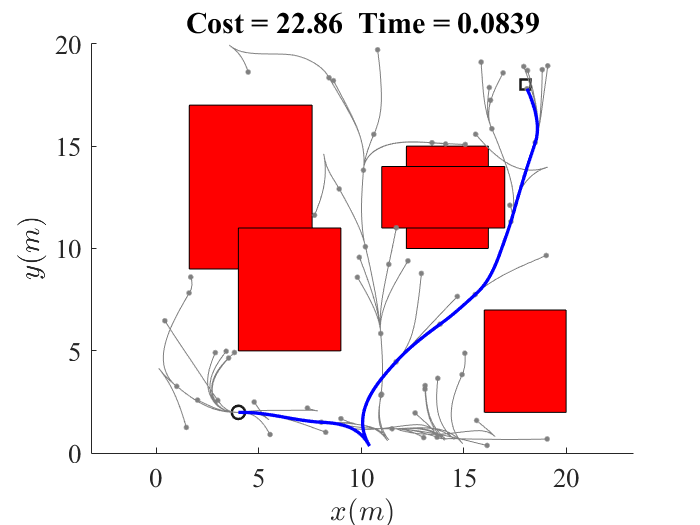}         
     \end{subfigure}
     \begin{subfigure}{1\columnwidth}
         \centering
         \includegraphics[width=0.49\columnwidth]{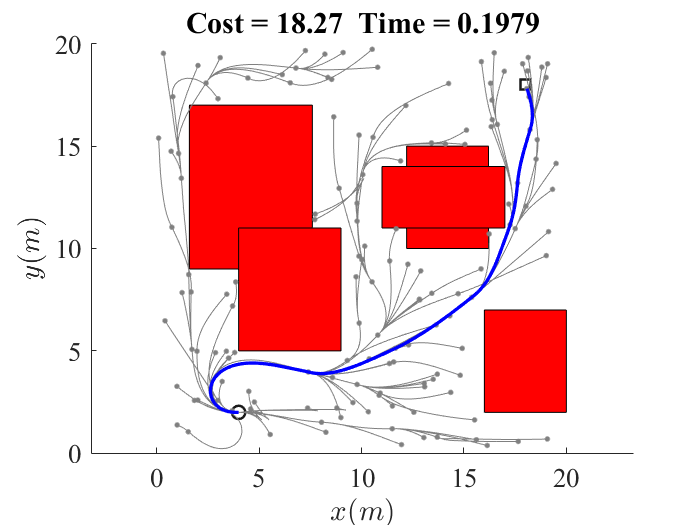}
         \includegraphics[width=0.49\columnwidth]{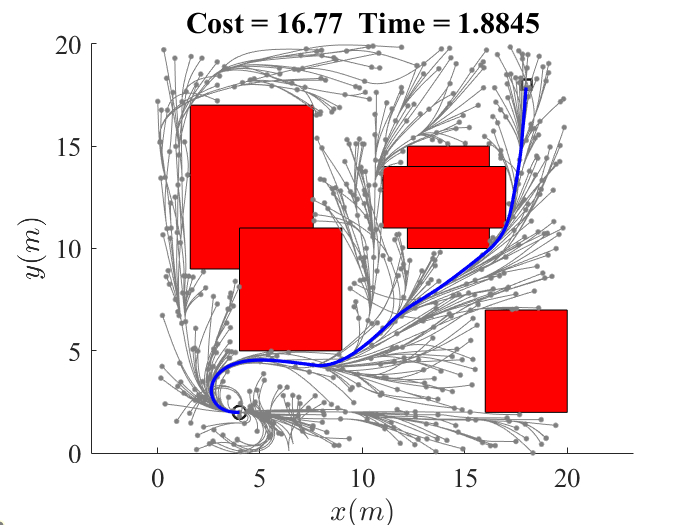}         
     \end{subfigure}
        \caption{FMT*PFF planning results for the car example. The blue line is the solution in the current tree.}
        \label{Treecar}
\end{figure}
\begin{figure}[!ht]
    \centering
    \includegraphics[width=0.5\columnwidth]{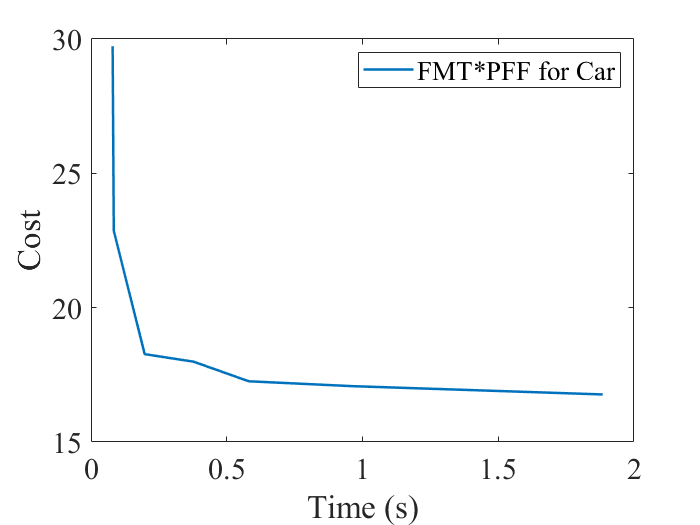}
        \caption{Performance of FMT*PFF with a neural network controller for the car example.}
        \label{CostCar}
\end{figure}

After training the neural network, we tested the neural network controller for randomly sampled initial states. 
Some example trajectories obtained using the neural network controller are shown in Figure~\ref{TestTraj}.
Since the goal of the neural network controller is to steer the system to reach the origin, one index is the error between the end position of the trajectories and the origin.
1000 trajectories corresponding to sampled novel initial states are generated using the neural network controller.
98\% of the resulting trajectory reached a 0.3 neighborhood of the origin. 
If this end position error is too large (greater than 0.3), the connection is not successful, and this edge will not be added to the tree. 
Note that FMT*PFF does not require exact steering, which makes it suitable for learning-based steering functions.
The final trajectory obtained from the FMT*PFF algorithm is smooth and satisfies the differential constraint, while the trajectory from \cite{zheng2021sampling} may have small gaps between edges.

The planning results of FMT*PFF using the neural network controller are shown in Figure~\ref{Treecar} and Figure~\ref{CostCar}. Figure~\ref{Treecar} shows the trees with a different number of samples and Figure~\ref{CostCar} gives the cost vs time performance. 

\begin{figure}[!ht]
    \centering
    \includegraphics[width=0.49\columnwidth]{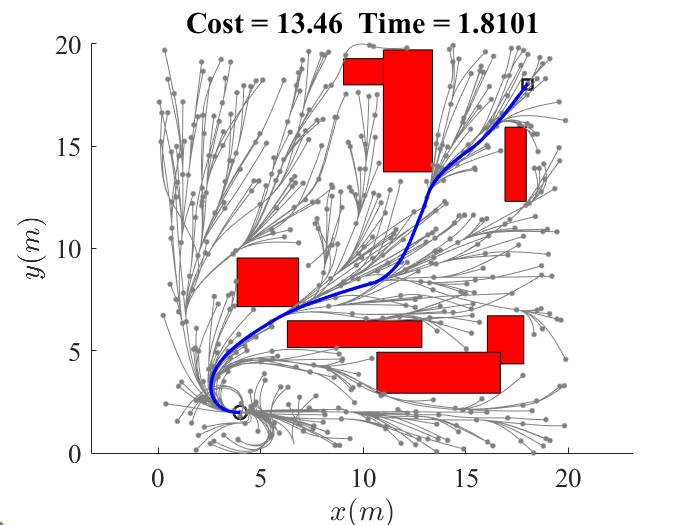} \includegraphics[width=0.49\columnwidth]{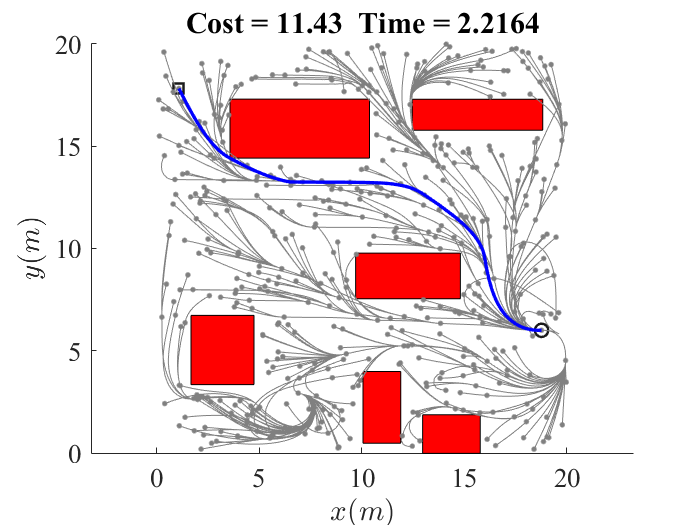}   
    \includegraphics[width=0.49\columnwidth]{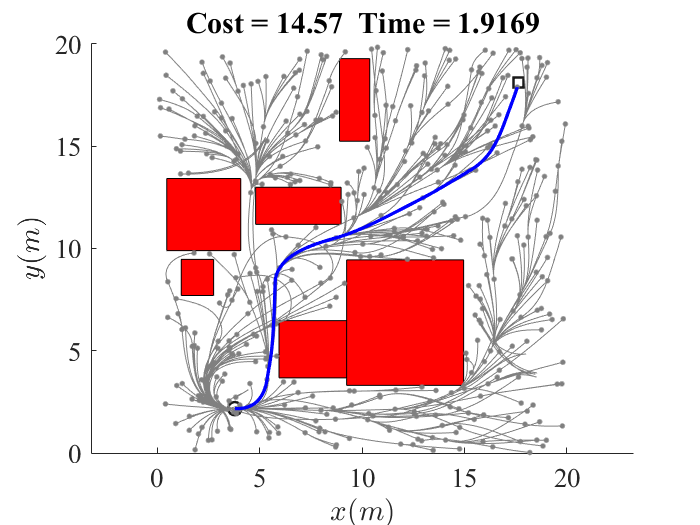}
    \includegraphics[width=0.49\columnwidth]{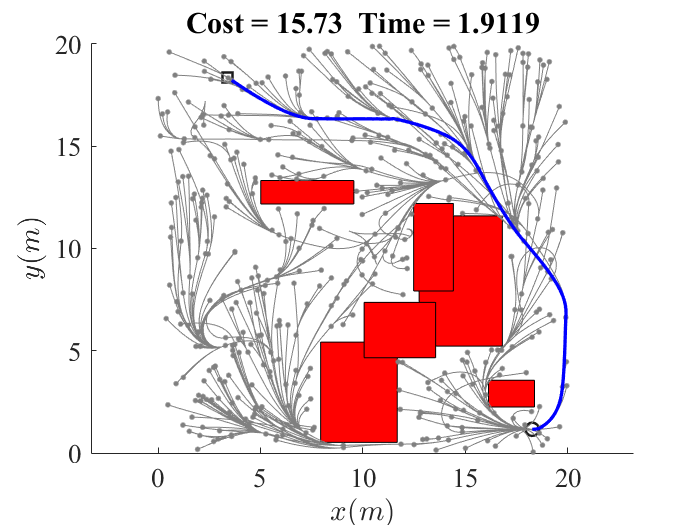} 
    \caption{Planning results for the car in different environments.}
    \label{CarEnvs}
\end{figure}

Finally, we use FMT*PFF to plan trajectories for the car model in various environment settings. 
We randomly sample the number, size, and location of the obstacles. We also vary the starting point and goal point of the car. 
The planning results are given in Figure~\ref{CarEnvs}.
The FMT*PFF algorithm finds dynamically feasible trajectories using the neural network controller.

\section{Conclusion}\label{sec:conclusion}

We propose the FMT*PFF algorithm for optimal kinodynamic motion planning for nonlinear systems.
The key idea is the use of a partial-final-state-free (PFF) optimal controller to reduce dimensionality and accelerate kinodynamic motion planning is introduced.
FMT*PFF planning in the reduced state space and has faster convergence.
By training a neural network model of the PFF optimal controller, FMT*PFF can plan trajectories for nonlinear systems in cluttered environments with nonconvex obstacles. 
FMT*PFF can deal with learning-based steering functions that can not achieve exact steering because no rewire is needed.
We show that FMT*PFF is efficient and generates dynamically feasible motion plans.
Through numerical simulations and comparison with previous works, FMT*PFF are shown to have better cost-time performance.

\bibliography{sample}

\end{document}